\def\year{2019}\relax
\documentclass[letterpaper]{article} 
\usepackage{times}
\usepackage{aaai19}  
\usepackage{helvet}
\usepackage{mathrsfs}
\usepackage{courier}
\usepackage{url}
\usepackage{graphicx}
\usepackage{multirow}
\usepackage[table]{xcolor}
\usepackage{amsmath}
\usepackage{amsfonts}
\usepackage[algo2e]{algorithm2e} 
\usepackage{amssymb}
\usepackage{algorithm,algpseudocode}
\usepackage{algorithmicx}
\usepackage{multirow}
\usepackage{bigstrut}

\usepackage{booktabs} 
\usepackage{colortbl} 
\usepackage{xcolor} 
\usepackage{xfrac}
\usepackage{amsmath}
\usepackage{caption} 
\begin{document}
%
\title{Optimizing Heuristics for Tableau-based OWL Reasoners}

\author{Razieh Mehri \\Concordia University\\Montr\'eal, Canada\\r\_mehrid@encs.concordia.ca\\\And Volker Haarslev\\ Concordia University\\Montr\'eal, Canada\\haarslev@cse.concordia.ca \And Hamidreza Chinaei\\ Nuance Communications\\hamid.chinaei@nuance.com}
\maketitle
\begin{abstract}
Optimization techniques play a significant role in improving description logic reasoners covering the Web Ontology Language (OWL). These techniques are essential to speed up these reasoners. Many of the optimization techniques are based on heuristic choices. Optimal heuristic selection makes these techniques more effective. The FaCT++ OWL reasoner and its Java version JFact implement an optimization technique called \textit{ToDo list} which is a substitute for a traditional top-down approach in tableau-based reasoners. The ToDo list mechanism allows one to arrange the order of applying different rules by giving each a priority. Compared to a top-down approach, the \textit{ToDo list} technique has a better control over the application of expansion rules. Learning the proper heuristic order for applying rules in \textit{ToDo list} will have a great impact on reasoning speed. We use a binary SVM technique to build our learning model. The model can help to choose ontology-specific order sets to speed up OWL reasoning. On average, our learning approach tested with 40 selected ontologies achieves a speedup of two orders of magnitude when compared to the worst rule ordering choice.
\end{abstract}

\section{Introduction and Related Work}

Description logic (DL) reasoners are employed to infer implicit information from ontologies by performing different reasoning tasks such as ontology consistency, class satisfiability, hierarchical classification of named classes, query related tasks, etc.
Followed by FaCT \cite{horrocks1998using}, FaCT++ \cite{tsarkov2006fact} 
was developed to provide more portability as well as new algorithms including a novel optimization technique called ToDo List \cite{tsarkov2007optimizing}. 

While former tableau reasoners apply a depth-first top-down approach (also called trace technique), FaCT++ uses a set of queues in order to determine the set of currently applicable rules \cite{schmidt1991attributive,DLhandbook,tsarkovdynamic}. The trace technique makes it difficult to implement inverse roles since they cause both up and down propagation in tableau models; while the queue mechanism used by FaCT++ is more flexible and better suited for more expressive OWL languages that offer inverse roles. Moreover, more heuristic optimizations are possible with ToDo list compared to the top-down approach, where  heuristics are only applicable at a local and not at the global level because only the current branch is stored in a trace tree \cite{tsarkov2005ordering}.

Learning to select the best heuristics in a reasoner is essential for the reasoning performance. The heuristic selection options also vary depending on the optimization techniques and their implementation in reasoners. For example, due to the JFact's configuration setting, the change of rule ordering in the ToDo List approach is done globally instead of locally (during the reasoning process); this may not be very applicable, and thus not useful for learning purposes due to the overhead runtime of the learning process in each tableau level.

Several research efforts have been conducted on learning different factors that have an impact on improving reasoning speed for reasoners. First, \cite{kang2012predicting} applied supervised machine learning to learn the feature behavior for predicting the performance of OWL reasoners. Later, \cite{sazonau2014predicting} came up with more relevant features for the reasoning performance of ontologies. The results show the hardness of predicting OWL reasoners' performance due to the complexity of ontologies. However, it confirms that considering the correlation of ontologies' features or metrics thereof improves the predictions. Another study investigates the usefulness of different features by categorizing them while expanding the works of the previously mentioned papers \cite{alaya2015makes}.

E-MaLeS is an automatic tuning framework for Automated Theorem Provers (ATP) that applies machine learning to improve ATP's performance by scheduling the suitable strategies for feature selection \cite{kuhlwein2013males}. 

When improving an optimization technique in reasoning process, the focus is the effect of specific factors such as changing the order of operations on reasoning tasks; therefore, the exact features and their tuning for the above systems cannot be generalized to our case. Features such as the number of operations and their correlated features may be more effective for our specific prediction problem, although this may need its own investigation.

In a related work, \cite{mehriapplying} focused on improving semantic branching for disjunctions and its effect on backjumping, which is an  optimization technique for backtracking. Even though semantic branching is guaranteed to avoid redundant search space occurring in traditional syntactic branching, machine learning techniques also help to further decrease redundant search space exploration. The task is possible by learning new orders for applying variables at each branching level. As a first step, they focused on propositional SAT testing. Their results show that machine learning speeds up JFact by one to two orders of magnitude \cite{mehriapplying}. Compared to their approach, in this paper, we improve the ToDo list optimization technique in JFact using machine learning and apply it to a complex description logic that is a syntactic variant of OWL and contains propositional logic as a very small subset. To the best of our knowledge this is a first machine learning approach for improving such a rule based optimization technique for OWL reasoners. Our approach is based on an independent systematic analysis of rule orderings to uncover ontology patterns and their associated features that are relevant for this specific optimization technique.

The remainder of the paper is as follows. In the background section, some description logic notation is introduced. Also, the ToDo list architecture is explained in detail. In the section on order labeling, the prerequisite steps for designing our learning model are presented. Our evaluation  results are presented in the section on experiments and in the last section a discussion and conclusion are given.

\section{Background}

\subsection{Description Logic}

Description logic is a logic-based knowledge representation language which is used to describe the knowledge of a domain and derive implicit knowledge from that is entailed. To build a knowledge base, description logic uses the following terminologies:

\begin{itemize}
\item Concept describes a class of objects which share similar characteristics, e.g., the concept $\mathit{Man}$ contains all individuals who are a man.
\item Role is a binary relationship between individuals or data values, e.g., role $\mathit{hasChild}$ could be a relationship between two individuals where one is an instance of the concept $\mathit{Parent}$ and the other is an instance of the concept $\mathit{Child}$.
\item Individual stands for an object or an instance of a class, e.g., $\mathit{Josh}$ could be an instance of the concept $\mathit{Parent}$ meaning $\mathit{Josh}$ is a parent.
\end{itemize}

Syntax and semantics of $\mathcal{SHOIQ}$, which is a syntactic variant of OWL, are given in Table \ref{tab:Infer}, using an interpretation $\mathcal{I}=(\Delta ^ \mathcal{I},\cdot^\mathcal{I})$ where $\Delta ^ \mathcal{I}$ is the non-empty domain and $\cdot^\mathcal{I}$ the  interpretation function. The axiom $\mathrm{tr}(R)$ declares a role $R$ as transitive. The concept $\top$ ($\bot$) is an abbreviation for  $\neg A\sqcup A$ ($\neg A\sqcap A$), where $A$ is a concept. A concept $C$ is satisfiable if $C^\mathcal{I} \neq \emptyset$. 

A $\mathcal{SHOIQ}$ knowledge base consists of a TBox and ABox. A TBox is a finite set of concept ($C \sqsubseteq D$) and possibly role inclusion ($R \sqsubseteq S$) axioms to represent general knowledge about a domain ($C,D$ concept expressions and $R,S$ are roles). A concept (role) inclusion axiom is satisfied by $\mathcal{I}$ if $C^{\mathcal{I}} \subseteq D^{\mathcal{I}}$ ($R^{\mathcal{I}} \subseteq S^{\mathcal{I}}$). A Tbox $\mathcal{T}$ is satisfiable if there exists an interpretation (or model) $\mathcal{I}$ that satisfies all inclusion axioms.

An ABox represents assertional knowledge for individuals of that domain. It is a finite set of instance (${a\!:\!C}$) and role assertions ($(a,b)\!:\!R$). An instance (role) assertion is satisfied if $a^{\mathcal{I}} \in C^{\mathcal{I}}$ ($(a^{\mathcal{I}},b^{\mathcal{I}}) \in R^{\mathcal{I}}$). An Abox $\mathcal{A}$ is satisfiable w.r.t\ $\mathcal{T}$ if there exists an interpretation $\mathcal{I}$ that satisfies $\mathcal{T}$ and all assertions in $\mathcal{A}$.


For example, consider the concept expression:\\
\centerline{$\mathit{Female \;\sqcap \geq 2\, hasChild \sqcap\forall hasChild.Female}$}

that describes females which have at least two children and all of their children are females. Three parts are related through the conjunction operator. The first part includes individuals who belong to the concept $\mathit{Female}$, the second part describes individuals connected to at least two individuals via the $\mathit{hasChild}$ role and the third part describes individuals ($\forall$) that are connected via the $\mathit{hasChild}$ role to individuals of the $\mathit{Female}$ class.

\begin{table}[t]
\caption{Syntax and semantics of $\mathcal{SHOIQ}$}
\label{tab:Infer}
\begin{tabular}{|l|l|l|}
\hline 
\textbf{Operator}\hspace*{-1mm} & \textbf{Syntax}\hspace*{-1mm} & \textbf{Semantics}\\
\hline 
\hline 
Concept  & $A$  & $A^{\mathcal{I}}\subseteq\Delta^{\mathcal{I}}$, $A$ is atomic \\
Negation  & $\neg C$  & $\Delta^{\mathcal{I}}\setminus C^{\mathcal{I}}$\\
Conjunct.  & $C\sqcap D$  & $C^{\mathcal{I}}\cap D^{\mathcal{I}}$\\
Disjunct.  & {\small{}$C\sqcup D$}  & {\small{}$C^{\mathcal{I}}\cup D^{\mathcal{I}}$}\\
Universal  & $\forall R.C$  & $\{x \,|\, \forall y\!:\! (x,y) \!\in\! R^{\mathcal{I}}\Rightarrow y \!\in\! C^{\mathcal{I}}\}$\hspace*{-3mm}\\
Existential & $\exists R.C$  & $\{x \,|\, \exists y\!:\! (x,y) \!\in\! R^{\mathcal{I}} \wedge  y \!\in\! C^{\mathcal{I}}\}$\\
At Least & $\ge n R.C$  & $\sharp Q(x,R,C) \ge n$\\
At Most & $\le n R.C$  & $\sharp Q(x,R,C) \le n$\\
\hline 
nominal  & $\{o\}$  & $\sharp{\{o\}^{\mathcal{I}}}=1$ \\
\hline 
Role  & $R$  & $R^{\mathcal{I}}\subseteq\Delta^{\mathcal{I}}\times\Delta^{\mathcal{I}}$\\
\hline
Inverse  & $R^{-}$  & $(R^{-})^{\mathcal{I}} \!=\! \{(y,x)\,|\,(x,y)\in\! R^{\mathcal{I}}\}$\\
Transitive & $\mathrm{tr}(R)$  & $R^{\mathcal{I}}=(R^{\mathcal{I}})^{+}$\\
\hline 
\end{tabular}
$Q(x,R,C) = \{y \,|\, (x,y) \in R^{\mathcal{I}} \wedge  y \!\in\! C^{\mathcal{I}}\} $
\end{table}


\subsubsection{Concept Inference Services}

\begin{itemize}
\item Satisfiability: A concept $C$ is said to be satisfiable considering the TBox $\mathcal{T}$ if and only if there is a model $\mathcal{I}$ 
in $\mathcal{T}$ where $C^\mathcal{I}$ is not empty or $\mathcal{T}\not\models C\equiv \bot$

\item Subsumption: The concept $C$ is subsumed by concept $D$  considering the TBox $\mathcal{T}$ if and only if $\mathcal{T}\models C\sqsubseteq
D$ or $C^\mathcal{I}\subseteq D^\mathcal{I}$ holds for every model $\mathcal{I}$ of $\mathcal{T}$.

\item Equivalence: Two concepts $C$ and $D$ are equivalent considering the TBox $\mathcal{T}$ if and only if $\mathcal{T}\models C\equiv D$ or $C^\mathcal{I}\equiv  D^\mathcal{I}$ holds for every model $\mathcal{I}$ of $\mathcal{T}$ or $C \sqsubseteq D$ and $D \sqsubseteq C$.

\item Disjointness: Two concepts $C$ and $D$ are disjoint considering the TBox $\mathcal{T}$ if and only if $\mathcal{T}\models C\sqcap
D\equiv \bot$ or $C^\mathcal{I}\cap D^\mathcal{I}=\emptyset$ holds for every model $\mathcal{I}$ of $\mathcal{T}$.
\end{itemize}
It is worth to mention that the above tasks can also be translated to each other:
\begin{itemize}
\item Unsatisfiability/Subsumption: $C$ is unsatisfiable if and only if $C$ is subsumed by $\bot$.
\item Subsumption/Disjointness: $D$ subsumes $C$ if and only if $\neg D$ and $C$ are disjoint.
\item Disjointness/Equivalence: Two concepts $C$ and $D$ are disjoint if and only if $C\sqcap D$
is equivalent to $\bot$.
\item Equivalence/Unsatisfiability: $C$ and $D$ are equivalent if and only if $(C\sqcap\neg D)\sqcup (D\sqcap\neg C)$ is not satisfiable.
\end{itemize}

\subsubsection{Completion Graph} A Completion Graph $G = \langle V,E,\mathcal{L} \rangle$ is a directed graph. Each node $x \in V$ is labeled with a set of concepts $\mathcal{L}(x)$. Each edge $(x,y) \in E$ is labelled with a set of roles $\mathcal{L}(x,y)$. A graph has a clash if a node label $\mathcal{L}(x)$ contains $\{\neg A,A\}$ for a concept $A$. A Graph is complete when no more expansion rules can be applied. 

\subsubsection{Tableau-based Reasoning}
Tableau-based reasoning was first introduced for the $\mathcal{ALC}$ language but then extended for many concept languages. The algorithm's structure is based on completion graphs. We assume that all concept expressions are in negation normal form.
\\
The completion graph for $\mathcal{ALC}$, which is a core subset of $\mathcal{SHOIQ}$, is expanded using tableau completion rules shown in Table~\ref{tab:Tableau}, until no rule can be applied, i.e., the result is a complete graph and it might contains a clash if $\{\neg A,A\}\subseteq \mathcal{L}(x)$ for some node $x$.

\begin{table}
\centering
\caption{Tableau completion rules for $\mathcal{ALC}$}
\begin{tabular}{|ll|}
\hline
$\sqcap$-\textbf{Rule} &\textbf{If} $(C\sqcap D)\in \mathcal{L}(x)$ and $\{C, D\}\not\subseteq \mathcal{L}(x)$  \\
         & \textbf{then} set $\mathcal{L}(x)=\mathcal{L}(x)\cup\{C, D\}$ \\
         
$\sqcup$-\textbf{Rule} & \textbf{If} $(C\sqcup D)\in \mathcal{L}(x)$ and $\{C, D\}\cap \mathcal{L}(x)=\emptyset$ \\
         &\textbf{then} \\
         & set $\mathcal{L}(x)=\mathcal{L}(x)\cup \{E\}$ with $E\in \{C, D\}$ \\
         
$\forall$-\textbf{Rule} &\textbf{If} $\forall R.C \in \mathcal{L}(x)$ and there is a node $y$ \\
& with $R\in \mathcal{L}(x,y)$ and $C\not\in\mathcal{L}(y)$\\
         &\textbf{then} set  $\mathcal{L}(y)=\mathcal{L}(y)\cup \{C\}$  \\
         
$\exists$-\textbf{Rule} &\textbf{If} $\exists R.C \in \mathcal{L}(x)$ and there is not any node $x$ \\
& with $R\in \mathcal{L}(x,y)$ and $C\in\mathcal{L}(y)$\\
         & \textbf{then} \\
         & set $\mathcal{L}(x,y)=\mathcal{L}(x,y)\cup\{R\}$ and \\
         & $\mathcal{L}(y)=\mathcal{L}(y)\cup \{C\}$ where $y$ is a new node   \\
\hline
\end{tabular}
\label{tab:Tableau}
\end{table}

\subsection{ToDo List Architecture}

Performing reasoning tasks can be costly in tableau-based systems due to non-deterministic expansions. To reduce space and time cost,  optimization techniques such as ordering of expansion rules are used to decide what rules to expand first. 

In the traditional top-down approach using the trace technique, the  $\exists$-rule has the lowest priority; therefore, no unnecessary graph expansion happens until there are no more rules to apply except the $\exists$-rule. Additionally, the trace technique only considers the local trace and not a fully expanded tree since it removes the already traced branches in order to save space. Therefore, it is difficult to apply the trace technique in the presence of inverse roles. The reason is that not only the successors but also the predecessors will be involved and must be expanded recursively if needed while the approach only keeps hold of local expansions. Moreover, traditional tableau-based systems apply non-deterministic rules such as $\sqcup$ earlier than any other rules which increases the potential search space. 

To overcome the above restrictions, ToDo list does not only store the entries in a particular order but also creates different queues based on the priority of different rule. Therefore, every time an entry is selected  from the ToDo list, it is chosen from the queue with the highest priority. The process ends when all queues are empty or a clash is detected. If all queues have the same priority and only $\exists$ has the lowest priority; then the ToDo list works the same way as the traditional top-down trace technique \cite{tsarkov2005ordering,knublauch2004protege}.

\section{Order Labeling}

The default order for rules in FaCT++ is based on the exposure to a limited number of test ontologies \cite{tsarkov2005ordering}. Yet, the effect of different orderings varies case to case depending on the features of ontologies and cannot be generalized for all ontologies in these reasoners. Moreover, \cite{tsarkov2005ordering} already concluded that there is no universal strategy to ensure the best ordering heuristics of rules in ToDo list for all types of ontologies. Hence, the best scenario could be an application of a learning strategy that considers the features of an ontology in order to predict its best order.

To tackle this as a supervised machine learning problem, the labels (classes) need to be identified. Here, each label is considered as an order set that defines the priority of rules in ToDo list. Having more than one order set, this is a multi-class classification problem with the number of classes equal to the number of possible order sets. Hence, for each input ontology (OWL file), its label is an assigned order set that gives the highest speed solution to the reasoner for that ontology. For the rest of this paper, the words ``label" and ``class" are used interchangeably.

JFact \cite{palmisano2015jfact}, a Java port of the FaCT++ has a configuration setting that allows the modification of rules' ordering in ToDo list \cite{tsarkov2006fact}. To do this, a string of characters is defined where each character specifies a rule. This string is identified as ``IAOEFLG" in which ``I'' stands for a concept  \textit{Id} number, ``A'' for conjunction ($\sqcap$), ``O'' for disjunction ($\sqcup$), ``E'' for existential quantifier ($\exists$), ``F'' for universal quantifier ($\forall$), ``L'' for less than or equal ($\leqslant$) and ``G'' for greater than or equal ($\geqslant$) rules, respectively. Each of these characters is assigned to a number (with $0$ being the highest priority) indicating the associated rule priority, e.g., $0612354$ specifies that \textit{Id} has the highest priority over other operations, then $\sqcup$ is the second priority, and so on.

JFact, was indented to have the same functionality as FaCT++ but it is believed to not have the exact same implementation and performance as FaCT++ \cite{gonccalves2013empirical}. However, the JFact default configuration, which was set to 1263005 by its developers, is deemed to be based on the FaCT++ code \cite{tsarkov2006fact} but such a configuration might not be adequate for currently available ontologies.

The number of all possible order sets obtained from assigning numbers to the characters is $7^7$ (considering that some rules can have the same priorities as well). In order to make the problem practical and applicable to the reasoner, the number of order sets needs to be reduced to only a few reasonable and effective order sets. Therefore, before prioritizing, some of the rules are combined based on how they affect the reasoning process. \textit{Id} is combined with $\sqcap$, $\geqslant$ is combined with $\exists$, and finally $\leqslant$ with $\sqcup$. This leads us to define only 7 order sets as shown in Table~\ref{tab:order} (rules in the same cells have the same priority). \textit{Id} and $\sqcap$ are ignored and given the first priority because they do not cause non-determinism or expansion, i.e., it is believed that their priority seldom has any effect on the reasoning time.
\begin{table}
\large
\centering
\caption{Labels for order sets}
\label{tab:order}
\begin{tabular}{c|c|c|c|c|c|}
\multirow{2}{*}{Label}&\multirow{2}{*}{Order} & \multicolumn{4}{c}{Priority} \\
\cline{3-6}
                          &      & 1     & 2     & 3     & 4    \\
                                 \hline \hline
1& 012312 & $\sqcap$& $\leqslant$,$\sqcup$&$\geqslant$,$\exists$&$\forall$\\ 
 \hline  
2& 013213 & $\sqcap$&$\leqslant$,$\sqcup$&$\forall$&$\geqslant$,$\exists$\\
  \hline
3& 000000 & $\sqcap$,$\leqslant$,$\sqcup$,$\geqslant$,$\exists$,$\forall$& NA & NA & NA \\
  \hline
4& 032132 & $\sqcap$&$\geqslant$,$\exists$&$\forall$&$\leqslant$,$\sqcup$ \\
  \hline
5& 031231 & $\sqcap$&$\forall$&$\geqslant$,$\exists$&$\leqslant$,$\sqcup$  \\
  \hline
6& 021321 & $\sqcap$&$\geqslant$,$\exists$&$\leqslant$,$\sqcup$&$\forall$\\ 
  \hline
7& 023123 & $\sqcap$&$\forall$&$\leqslant$,$\sqcup$&$\geqslant$,$\exists$\\\hline
\end{tabular}
\end{table}

Earlier, a multi-classification learner was suggested as a solution to learn the best order among the 7 order sets for each input ontology. With the small number of training data available for this task, the number of labels (7) is pretty high and leads to only 50\% accuracy; therefore, the alternative solution is to consider each order set as a separate binary classification problem and predict if for an input ontology, that is a proper order set or not. More details on this approach are given in the Experiment Section. 

\subsection{Feature Engineering and Selection}
The importance of using good features is an essential topic among machine learning researchers and it directly has an effect on the performance of the built model. Feature engineering is a process of creating proper features from the current features. These feature extractions make a significant impact on the accuracy of the machine learning model. 

One feature engineering technique is to consider combining the features, e.g., adding the number of disjunctions and conjunctions together and consider them as one feature instead of having them as two separate features. However, based on the domain and the importance of features, combining features can have a positive or negative effect. Another feature engineering technique which has been used for data slicing in topics such as data mining tasks is bucketization. This can be used for converting features with range values into buckets \cite{Shekhar:2018}.
Moreover, feature selection is used to reduce the dimensionality of feature space and thus increase the accuracy of the model. 

Feature selection techniques that is used in our case is mutual information that calculates a score measure for how much a feature can contribute to a correct predicted class \cite{cover1991entropy}. 
\begin{equation}
MI(X,Y)={\sum_{x\in X}^{}}{\sum_{y\in Y}^{}} p(x,y) \log _{} {\frac{p(x,y)}{p(x)p(y)}}
\end{equation}
$X$ stand for features and $Y$ stand for classes. With mutual information, one can choose the $k$ top useful features for their model.

Principal component analysis (PCA) is another feature reduction technique that considers the co-relation between the features and their variance. PCA is used to reduce the feature size of the selected features while minimizing the information or variance loss.

\subsubsection{Ontology Features} Ontology features are categorized into four categories \cite{alaya2015towards}:

\begin{enumerate}
\item Ontology size which contains the basic measures of an ontology's signature and axioms, e.g., number of named classes, object properties, logical axioms, etc. 
\item Ontology expressively which describes the DL family name of an ontology. 
\item Structural feature of an ontology which contains the features related to hierarchy, richness and cohesion of the ontology, e.g., maximum depth of the ontology, number of subclasses, subproperties, ratio of classes having attributes, etc. 
\item Syntactical feature of an ontology which contains different ratios and frequencies related to classes (and their different types), properties, individuals, axioms and constructors.
\end{enumerate}

We consider 48 features which are discussed in the following. 
The basic features from the 48 features used in this work are mentioned in Table~\ref{tab:features}.
\begin{table}[t]
\renewcommand{\arraystretch}{1.2}
\caption{Basic features for ontologies}
\label{tab:features}
\centering
\begin{tabular}{c|c}
 \hline
   & Ontology Features \\ 
 \hline\hline
 1 & Number of Existential Value Restriction of Roles  \\ 
 \hline
 2 & Number of Universal Value Restriction of Roles  \\
 \hline
 3 & Number of Classes \\
 \hline
 4 & Number of Conjunction Groups \\
 \hline
 5 & Number of Disjunction Groups\\ 
 \hline
 6 & Number of Disjoint Classes \\ 
 \hline
 7 & Number of Object Properties \\ 
 \hline
 8 & Number of Inverse Object Properties \\ 
 \hline
 9 & Number of Nominals \\ 
 \hline
 10 & Number of Instances \\ 
  \hline
 11 & Number of Role Assertions\\ 
  \hline
 13 & Number of Min Cardinalities \\
 \hline
 14 & Number of Max Cardinalities \\
 \hline
  15 & Number of Subclasses \\
 \hline
  16 & Number of Equivalent Classes \\
  \hline
  17 & Number of Sub Object Properties \\ 
 \hline
  18 & Number of Domains \\
  \hline
  19 & Number of Ranges \\ 
  \hline
  20 & Number of Data Properties \\ 
  \hline
  21 & Number of Data Properties Assertions 
\end{tabular}
\end{table}
In addition to the selected features, we also added the number of different types of object properties including functional, transitive, symmetric, inverse functional object properties. 
\\
Moreover, the ratio of TBox axioms, ABox axioms and RBox axioms  to the total number logical axioms are added as 3 different features.
\\
To add more useful features, we also added the ratio of each rule category to the whole number of rules.
These ratio features are:
\begin{align} \label{ratio1}
\mathit{Ratio_{\leqslant+\forall}=\frac{No_\leqslant+No_\forall}{No_\leqslant+No_\forall+No_\sqcup+No_\sqcap+No_\geqslant+No_\exists}}
\end{align}
\begin{align} \label{ratio2}
\mathit{Ratio_{\geqslant+\exists}=\frac{No_\geqslant+No_\exists}{No_\leqslant+No_\forall+No_\sqcup+No_\sqcap+No_\geqslant+No_\exists}}
\end{align}
\begin{align} \label{ratio3}
Ratio_{\sqcup}=\frac{No_\sqcup}{No_\leqslant+No_\forall+No_\sqcup+No_\sqcap+No_\geqslant+No_\exists}
\end{align}
\begin{align} \label{ratio4}
Ratio_{\sqcap}=\frac{No_\sqcap}{No_\leqslant+No_\forall+No_\sqcup+No_\sqcap+No_\geqslant+No_\exists}
\end{align}

Another features that we found to be useful in our case due to the importance of instances, is the Average Population which indicates the ratio of instances to the number of classes in an ontology \cite{tartir2005ontoqa}.

Obtaining some of the features such as depth of hierarchy is very expensive especially for big size ontologies. That also may not be very substantial for the current task that is the impact of changing the rule order in reasoning tasks. 

The reasoning task used in this paper is the hierarchical classification of an ontology which computes the hierarchical relation between classes of the ontology. The term classification of ontologies is different from the term classification used for machine learning and to avoid the confusion we refer to the former as hierarchical classification of ontologies to avoid the confusion in this paper.

Since the reasoner's task is to find the hierarchical relations in ontologies, to obtain more knowledge on the deeper level of ontologies, we also considered the number of occurrences of each rule that $\mathit{SubClassOf}$ and $\mathit{EquivalentClasses}$ expressions in OWL files start with. These patterns for $\sqsubseteq$ are:
\begin{equation}
C \sqsubseteq (\sqcup (\ldots) ) , C \sqsubseteq (\leqslant (\ldots) )
\label{eqn:patterns1}
\end{equation}
\begin{equation}
C \sqsubseteq (\sqcap  (\ldots) ) 
\label{eqn:patterns2}
\end{equation}
\begin{equation}
C \sqsubseteq (\forall  (\ldots) )
\label{eqn:patterns3}
\end{equation}
\begin{equation}
C \sqsubseteq (\exists  (\ldots) ) , C \sqsubseteq (\geqslant (\ldots) )
\label{eqn:patterns4}
\end{equation}
We apply the same patterns for $\equiv$.

In addition to that, the number of occurrences of each rules in the parenthesis of each of the above patterns will also be considered as separate features. For example, for the pattern $C \sqsubseteq (\sqcup (\ldots) )$, we calculated the number of occurrences of all $\forall$s inside the parenthesis and saved it as a feature. 
\begin{equation}
C \sqsubseteq (\sqcup (\ldots ,\underline{\forall R.C_1},\ldots,\sqcup( C_2,\underline{\forall R.C_3},\ldots),\ldots ))
\end{equation}

The same is applied for other rules for each of the patterns in equations \eqref{eqn:patterns1}, \eqref{eqn:patterns2}, \eqref{eqn:patterns3}, and \eqref{eqn:patterns4}. 

\section{Experiments}

For this experiment, we collected ontologies from the OWL Reasoner Evaluation (ORE) 2014 \cite{bail2014summary}. Although the ORE 2014 corpus comprises of a large number of ontologies, not all of them can be considered in our dataset. The extracted ontologies from the ORE 2014 library are the ones that contain at least one of the rules in each of the prioritized categories of operands of Table~\ref{tab:order}. The reason is that the ToDo list technique is specifically designed to determine the order of some rules and if some of the selected ontologies do not share the same criteria for the label definitions. Furthermore this study does not focus on any machine learning problems where required features are missing. With this strategy, we collect around  1840 OWL files from the ORE 2014 repository.

Furthermore, another decision is required to examine if an ontology among 1840 OWL files is an appropriate indication of the training data or not. To make this decision, the reasoning task (hierarchical classification of ontologies) is performed on the 1840 data and the average of three runtimes (in order to obtain reliable runtimes) is calculated for each data on each order set of Table~\ref{tab:order}. Then two further conditions are implied.


First, if for an ontology the difference between the maximum and minimum runtimes with all the 7 order sets is less than 2 seconds, it will be excluded. For example, if classifying of an ontology takes 3347, 3357, 4537, 2851, 3066, 3408, 2951 milliseconds with the 7 order sets, respectively. In this example, the minimum and maximum runtimes are 2851 (belongs to order set 4) and 4537 (belongs to order set 3) and the difference between them is 1686 milliseconds. The difference of less than 2 seconds could be caused by overhead runtime. To make sure this is the case, we run it several times and in all runtimes the configurations with the biggest and smallest runtimes are not the same as the previous runtimes, i.e., the difference is indeed caused overhead runtime. For ontologies with the same condition this indicates that the rule ordering does not have any effects on the ontology (those ontologies could also be very simple and small). To keep the homogeneity conditions for our training data (and test data) those ontologies are excluded.  

Second, the training data that leads to timeouts for all 7 orders are excluded too. The timeout considered for this experiment is 500,000 milliseconds (8.33 minutes). Since obtaining the exact time for timeout cases are very expensive, we do not associate them with any numbers. Therefore, again it is not obvious whether the ontology is affected by any rule orderings.

The above conditions lead to 159 OWL files for our training data where we dedicate 25\% to test data. 

As mentioned previously, if each order set of Table~\ref{tab:order} is observed as one class this leads to a multi-classification model with a high prediction error rate. This is due to the high numbers of labels without large training data. Therefore, we take another approach and convert the problem into 7 binary classification problems (each order set as an independent machine learning problem) with two labels for each problem. The label ``Good" for an order set indicates that the order set solves the ontology (performs hierarchical classification) in less than a threshold value and the label ``Bad" indicates that the order set solves an ontology in more than a threshold value (the ``Bad" labels also includes timeout cases). The threshold in this experiment is obtained by exploiting the data distribution for all of the configurations (order sets).

\begin{table}[t]
\centering
\caption{Threshold (mean+std) for 7 configurations (std stands for standard deviation)}
\label{tab:stdmean}
\begin{tabular}{|c|c|c|c|}
\hline
Config & std & mean & std + mean \\
\hline
1 & 94811 & 56323 & 151134 \\
2 & 65705 & 30427 & 96132 \\
3 & 86221 & 42644 & 128865 \\
4 & 74743 & 33428 & 108171 \\
5 & 69883 & 33822 & 103705 \\
6 & 86980 & 45085 & 132065 \\
7 & 69312 & 30930 & 100242 \\
\hline
\end{tabular}
\end{table}

Table~\ref{tab:stdmean} shows the mean and standard deviation (std) for the training data in each configuration in milliseconds. The maximum value is 151134 which belongs to configuration 1 and the minimum value is 96132 which belongs to configuration 2. The average of mean plus standard deviation from all of the configurations is considered as the threshold value. This value is 117,187 milliseconds (around 2 minutes). The timeout cases are excluded while examining the data distributions.

To build the binary classification models of the configurations, we use SVM technique with 10-fold cross-validation. We conducted a grid search to select the best models by choosing the best parameters such as the numbers of PCA components. The optimal kernel used for all configurations except the second one is linear. Radial Basis Function (RBF) kernel outperforms linear by 10\% for configuration 2.


\begin{table}[t]
\centering
\caption{Cross-validation accuracy of the 7 configurations after applying Standardization, PCA (combined with 40 selected features from mutual information based feature selection) and SVM}
\label{tab:configaccuracypar}
\begin{tabular}{|c|c|c|}
\hline
Config & Accuracy & Priority for ``Good" labels \\
\hline
1 & 77\% & 5 \\
2 & 85\% & 1 \\
3 & 80\% & 4 \\
4 & 82\% & 3 \\
5 & 76\% & 6 \\
6 & 71\% & 7 \\
7 & 83\% & 2 \\
\hline
\end{tabular}
\end{table}

Table~\ref{tab:configaccuracypar} shows the cross-validation accuracy of all configurations after applying standardization and the combination of PCA with mutual information feature selection technique that selects 40 features with the highest scores.

\begin{table}[t]
\caption{F-score for all 40 tests chosen from ORE 2014}
\label{tab:fscore}
\centering
\begin{tabular}{|c|c|}
\hline
Config & F-score \\
\hline
1 & 78\%  \\
2 & 96\%  \\
3 & 87\%  \\
4 & 89\%  \\
5 & 75\%  \\
6 & 70\%  \\
7 & 100\%  \\
\hline
\end{tabular}
\end{table}

Finally, to assess the impact of the built models on JFact, 40 unseen samples from ORE 2014 competition are selected randomly (that also follow the two earlier-mentioned conditions for training data) and are evaluated by F1-score (Table~\ref{tab:fscore}). 

Table~\ref{tab:samples} shows the JFact runtime for 11 of 40 samples with different order sets. The last column shows the JFact runtime for the standard configuration. This indicates that compared to the standard JFact configuration that leads to timeout for samples 6 and 11, our learned based reasoner runs by 2 and 3 order of magnitudes faster, respectively. Even though in very few cases standard JFact usually surpass by only less than one order of magnitude it never outperforms our learned based reasoner with more than an order of magnitude and this small ratio also can easily be improved by increasing the accuracies of some configurations in our approach in our future work.

\begin{table*}[tp]
\caption{11 out of 40 tests chosen from ORE 2014 competition, $\mathit{TO}$ indicates a timeout and the selected configurations by the learned based reasoner are shown in bold, $^-$ and $^*$ indicate false negative (falsely predicted as ``Bad") and false positive (falsely predicted as ``Good"), respectively}
\label{tab:samples}
\centering
\begin{tabular}{|l|c|c|c|c|c|c|c||c|}
\hline
 Sample & Config 1 & Config 2 & Config 3 & Config 4 & Config 5 & Config 6 & Config 7 &JFact Standard Config\\
\hline
1 & $TO$ & $TO$ & $TO$ & \textbf{40146} & 39200 & 42831 & $TO$&40246\\
2 & $TO$ & \textbf{625} & 2050 & 903 & $TO$ & $TO$ & 607 & 881\\
3 & $TO$ & $TO$ & $TO$ & \textbf{15688} & 15253 & $19141^-$ & $TO$ & 15727\\
4 & 10296 & \textbf{9642} & 25926 & 8617 & 8761 & 8875 & 9451 & 8467\\
5 & 14272 & \textbf{13862} & 52538 & 15319 & 15508 & 16023 & 13672& 11240\\
6 & 2509 & \textbf{2529} & 2471 & $TO$ & $TO^*$ & $TO^*$ & 2498 & $TO$\\
7 & $4212^-$ & \textbf{4050} & $297164^*$ & $TO$ & 3808 & $TO$ & 3959 & 2701\\ 
8 & $TO$ & \textbf{417} & 1372 & 412 & 408 & 414 & 407 & 382\\
9 &  167748& \textbf{63112}& $TO^*$& 88423& 212917& 300729& 75658 & 85404\\
10 &  $TO$&$TO$&$TO$& \textbf{192741}& $198465^*$& $210961^*$& $TO^*$ & 188800\\
11 & $379^-$ & $373^-$ & 355 & $TO$ & $TO$  & $TO$ & \textbf{325} & $TO$\\
\hline
\end{tabular}
\end{table*} 

Moreover, Table~\ref{tab:speedup} shows the speedup factor of the learned order compared to the worst case ordering of rules, which is by average about two orders of magnitude faster. There were no negative speedup ratios meaning all of the 40 samples were performed better than the worst case scenario.

\begin{table}[t]
\caption{Speedup ratio of the selected order to the worst case}
\label{tab:speedup}
\centering
\begin{tabular}{|c|c|c|c|}
\hline
 &Min & Max & Average \\
\hline
Speedup Ratio Improvement & 1.29 & 1536.71 & 337.81 \\
\hline
\end{tabular}
\end{table}

\section{Discussion}

Binary classifiers built for each configuration (order set) have two labels (classes): ``Good" and ``Bad" indicating if an ontology can be classified in less than 2 minutes or more than 2 minutes including timeouts, respectively. This is very helpful for the samples that have timeouts or run in more than 2 minutes in some order sets and can choose another order set that runs in less than 2 minutes. For example, in Table \ref{tab:samples}, since for sample 1, the configurations 1,2,3 and 7 are classified as ``Bad" order sets, with the built model, JFact will choose the configuration 4 or 5 or 6 instead, i.e., it will run in less than 1 minute instead of more than 8 minutes.

The main strategy is to ignore the ``Bad" order sets and choose the ``Good" order sets. In this case, we end up with three situations: 

\begin{enumerate}
\item When there is only one ``Good" label among all the 7 configurations and others are ``Bad"; without hesitation, one chooses the ``Good" label. \item When there are multiple ``Good" labels, the learner will choose among the ``Good"s for the one with the highest accuracy (the priorities based on the cross-validation accuracies are given in Table~\ref{tab:configaccuracypar}). 
\item When all of the labels are predicted as ``Bad"; but the classifier will not indicate which one is timeout or less than 8 minutes. In this scenario, the best strategy is to choose the configuration with the lowest cross-validation accuracy in Table~\ref{tab:configaccuracypar}; presuming that it falsely detected as a ``Bad" label. However, only less than 10\% of the provided data with all ``Bad" labels contain both timeout and not timeout; thus, defining another label that is representing ``Bad" labels without timeout is not very efficient and decreases the accuracy of the learner. 
\end{enumerate}

The selected configurations for the samples are shown in bold in Table \ref{tab:samples}.  

Finally, if we consider the whole process of classifying ontologies in JFact this task is also accompanied with  pre-required reasoner tasks such as checking the consistency of ontologies. Depending on the size of nominals (or ABox), the consistency checking before the main hierarchical classification may take time (and even lead to a timeout). However, we believe these ontologies will also take almost the same amount of time for classification. The reason is that too many nominals indicate a big size ontology associated with more classes.
In order to cover all possible ontologies, we provided both training data and samples that contain ontologies with a big or small number, or no nominals.

\section{Conclusion and Future Work}

Although optimization techniques were designed to speed up the reasoning process for OWL reasoners, many of them are heuristic-based and could be modified case by case based on the type of ontologies. In this paper we focused on improving ToDo list that is a heuristic-based optimization technique of JFact reasoner. The order of rules in ToDo list is likely to have drastic impact on improving the performance of JFact. Thus, we learned a model that helps the reasoner to choose the sufficient heuristic for a given ontology and help them to run in less than 2 minutes and avoid the time out which is about 8 minutes.  

Although ontologies are very complex infrastructures, we believe by defining proper and correlated features, they can be classified for these optimization techniques of the reasoning process. In this paper, we presented a learned model that in several few cases outperforms and many other cases is close to the performance of the JFact standard configuration. Furthermore, the learned model outperforms the worst case ordering selection by 1 to 3 orders of magnitude. In our future work, we plan to improve our built model by determining better features that yield a more accurate model that is also able to detect the best time possible. Moreover, we yet have to discover what other heuristic optimization techniques can be improved by machine learning in JFact and other reasoners.
\bibliographystyle{aaai}
\bibliography{paper}
\appendix

\section{Appendix A: Definitions}

\subsection{Support Vector Machine}

Support Vector Machine (SVM) is a machine learning technique used for binary classification of data. Given data points ($(x_1,y_1),(x_2,y_2),\dots,(x_n,y_n)$ where $y_i$ is the class ($y_i$ is either class 1 or class 2) that $x_i$ belongs to, there are many lines that can be fitted to separate the data points based on the class that they belong to. The optimal hyperplane is the one that has the maximal margin or maximal distance between two data points of two classes. The closest data points of both classes are called supporting vectors. The hyperplane is the linear combination of supporting vectors multiplied by LaGrange multipliers.

Kernels were defined for SVM to support data that must be separated non-linearly. Three most popular kernels introduced for SVM are:  Linear, Gaussian (RBF) and Polynomial. Linear SVM is effective when data is larger and it simply separates the data by a line. Compared to the linear kernel, RBF and Poly are more effective since they are more flexible in separating the data. Besides, SVM is designed to handle high dimensional data that makes it a good choice for our case.

Although SVM is designed for binary classification, due to their efficiency, many tried to extend the SVM for more than 2 classes. Two popular methods are: One vs All (\textit{OvA}) and One vs One (\textit{OvO}). In \textit{OvA}, for each label, a binary classifier will be created where that label is considered as a positive label and others are negative labels; therefore, overall with $N$ labels, $N$ classifiers are built. In \textit{OvO}, for each pair of labels one classifier is created; therefore, $\frac{N(N-1)}{2}$ number of classifiers are built which is very expensive if $N$ is large \cite{van2016gensvm}.

\subsection{Principal Component Analysis}

PCA is an unsupervised learning technique for feature extraction. It reduces the dimension of data and defines a new set of dimensions based on the number of components. The components project the data into a direction, i.e., it maximizes the variance of data; in other words, spreading out the data. The first component has the greatest variance, the second component has the second greatest variance and so on. Standardization, which is an important practice before PCA, helps PCA to increase the variance in more than one component, i.e, makes use of important features. Standardization transforms the data in order to have a mean of zero and a standard deviation of one, i.e., the data points have the same scale for further transformations by classifiers. 

\section{Appendix B: Complete Table of Samples}
Table~\ref{tab:samples40} contains all 40 samples. It is submitted as supplemental material.

\begin{table*}[tp]
\caption{The 40 tests chosen from ORE 2014 competition, $\mathit{TO}$ indicates a timeout and the selected configurations by the learned based reasoner are shown in bold, $^-$ and $^*$ indicate false negative (falsely predicted as ``Bad") and false positive (falsely predicted as ``Good"), respectively}
\label{tab:samples40}
\begin{center}
\begin{tabular}{|l|c|c|c|c|c|c|c||c|}
\hline
 Sample & Config 1 & Config 2 & Config 3 & Config 4 & Config 5 & Config 6 & Config 7 &JFact Standard Configration\\
\hline
1 & $TO$ & $TO$ & $TO$ & \textbf{40146} & 39200 & 42831 & $TO$&40246\\
2 & $TO$ & \textbf{625} & 2050 & 903 & $TO$ & $TO$ & 607 & 881\\
3 & $TO$ & $TO$ & $TO$ & \textbf{15688} & 15253 & $19141^-$ & $TO$ & 15727\\
4 & 10296 & \textbf{9642} & 25926 & 8617 & 8761 & 8875 & 9451 & 8467\\
5 & 14272 & \textbf{13862} & 52538 & 15319 & 15508 & 16023 & 13672& 11240\\
6 & 2509 & \textbf{2529} & 2471 & $TO$ & $TO^*$ & $TO^*$ & 2498 & $TO$\\
7 & $4212^-$ & \textbf{4050} & $297164^*$ & $TO$ & 3808 & $TO$ & 3959 & 2701\\ 
8 & $TO$ & \textbf{417} & 1372 & 412 & 408 & 414 & 407 & 382\\
9 &  167748& \textbf{63112}& $TO^*$& 88423& 212917& 300729& 75658 & 85404\\
10 &  $TO$&$TO$&$TO$& \textbf{192741}& $198465^*$& $210961^*$& $TO^*$ & 188800\\
11 & $379^-$ & $373^-$ & 355 & $TO$ & $TO$  & $TO$ & \textbf{325} & $TO$\\
12 & $TO$& \textbf{557}& 936& 598& $TO$& $TO$& 557& 595 \\
13 & $TO$&  $TO$&  $TO$& \textbf{39500}& 38936& 41690& $TO$ & 40160\\
14 & $TO$& $TO$& $TO$& \textbf{36773}& 36602& $TO$& $TO$ & 35125\\
15 & $TO$& \textbf{14007}& 17900& 13916& $TO$& $TO$& 13882 & 13942\\
16 & $TO$& \textbf{6450}& 9905& 6647& $TO$& $TO$& 6556 & 8263\\
17 & $TO$& $TO$& $TO$& \textbf{42277}& 41877& 45745& $TO$ & 43638\\
18 & $TO$& \textbf{720}& 1171& 794& $867^-$& $TO$& 677 & 870\\
19 & $TO$&\textbf{773}& 2612& 856& 836& $TO^*$& 675 & 852\\
20 & $TO$&$TO$&$TO$& \textbf{15298}& 15626& $17728^-$ & $TO$ & 15269\\
21 & 2462& \textbf{2946}& 3599& 1553& 1536& 4586& 2639 & 1699\\
22 & $TO$& $518^-$& 475& 497& 469& 486& \textbf{476} & 589\\
23 & $718^-$& \textbf{678}& 3958& 1620& $1857^-$& $1932^-$& 664 & 1749\\
24 &  $TO$& \textbf{434}& 671& 417& $413^-$&  $TO$& 430 & 388\\
25 &  65579& \textbf{3432}& 18415& 2510& $5156^-$ & $25909^-$ & 3494 & 2734\\
26 &  $TO$& \textbf{457}& 471& 480& $TO^*$&$TO^*$& 457 & 456\\
27 &  $TO$&$TO$&$TO$& $\textbf{324682}^*$& $329102^*$& $350004^*$& $TO$ & 298338\\
28 & 230550 & \textbf{16057} & $178259^*$ & 3863 & 3879 & 27882 & 6575 & 3568\\
29 &  3717& \textbf{1058}& 3454& $1187^-$& 1497& 4050& 1083 & 1146\\
30 & $TO$ & \textbf{16105} & $TO^*$ & 2928 & $TO$ & $TO$ & 16679 & 2628\\
31 &  $7858^-$& $6982^-$& $99669^-$& $6436^-$& $6404^*-$& $5748^-$& \textbf{7048} & 6344\\
32 &  327919& \textbf{27042}& $264452^*$& 4357& 4397& 52241& 8343 & 3597\\
33 &  $TO$& \textbf{420}& 853& 436& $424^-$& $TO$& 415 & 443\\
34 &  $TO$& \textbf{545}& 574& $558^-$& $610^-$& $TO$& 553 & 581\\
35 &  3489& \textbf{919}& 2416& $931^-$& 1123& 3602& 910 & 848\\
36 &  $TO$& \textbf{449}& 791& 443& $402^-$& $TO$& 429 & 440\\
37 &  $TO^*$& \textbf{979}& 6022& 1714& $TO^*$&$TO^*$& 991 & 1753\\
38 &  574& \textbf{566}& 23380& 606& 586& 569& 581 & 615\\
39 &  91032& \textbf{96408}& $124938^*$& 2296& 2325& 2350& 93088 & 2405\\
40 &  17058& \textbf{3523}& 31232& $2172^-$& 2206& $17998^-$& 2469 & 1925\\

\hline
\end{tabular}
\end{center}
This table contains supplemental material and lists the results for all the tested 40 ontologies.
\end{table*} 

\end{document}